\documentclass[runningheads,a4paper]{llncs}

\usepackage{amssymb,amsmath}
\usepackage{graphicx,color}
\usepackage{rotating}
\usepackage{url}
\newcommand{\keywords}[1]{\par\addvspace\baselineskip
\noindent\keywordname\enspace\ignorespaces#1}

\begin{document}

\mainmatter  

\title{Interlinked Convolutional Neural Networks for Face Parsing}

\titlerunning{iCNN for Face Parsing}

%
%
\author{Yisu Zhou, Xiaolin Hu, Bo Zhang}
\authorrunning{Y. Zhou, X. Hu, B. Zhang}

\institute{State Key Laboratory of Intelligent Technology and Systems\\
Tsinghua National Laboratory for Information Science and Technology (TNList)\\
Department of Computer Science and Technology\\
Tsinghua University, Beijing 100084, China\\
}

%
%

\toctitle{Lecture Notes in Computer Science}
\tocauthor{Authors' Instructions}
\maketitle

\begin{abstract}
Face parsing is a basic task in face image analysis. It amounts to labeling each pixel with appropriate facial parts such as eyes and nose. In the paper, we present a interlinked convolutional neural network (iCNN) for solving this problem in an end-to-end fashion. It consists of multiple convolutional neural networks (CNNs) taking input in different scales. A special interlinking layer is designed to allow the CNNs to exchange information, enabling them  to integrate local and contextual information efficiently. The hallmark of iCNN is the extensive use of downsampling and upsampling in the interlinking layers, while traditional CNNs usually uses downsampling only. A two-stage pipeline is proposed for face parsing and both stages use iCNN. The first stage localizes facial parts in the size-reduced image and the second stage labels the pixels in the identified facial parts in the original image. On a benchmark dataset we have obtained better results than the state-of-the-art methods.
\end{abstract}

\section{Introduction}

The task of image parsing (or scene labeling) is to label each pixel in an image to different classes, e.g., person, sky, street and so on \cite{Tu05}. This task is very challenging as it implies jointly solving detection, segmentation and recognition problems \cite{Tu05}. In recent years, many deep learning methods have been proposed for solving this problem including recursive neural network \cite{Socher11}, multiscale convolutional neural network (CNN) \cite{Farabet13} and recurrent CNN \cite{Pinheiro14}. To label a pixel with an appropriate category, we must take into account the information of its surrounding pixels, because isolated pixels do not exhibit any category information. To make use of the context, deep learning models usually integrate multiscale information of the input. Farabet et al. \cite{Farabet13} extract multi-scale features from image pyramid using CNN. Pinheiro et al. \cite{Pinheiro14} solve the problem using recurrent CNN, where the coarser image is processed by a CNN first, then the CNN repeatedly takes its own output and the finer image as the joint input and proceeds. Socher et al. \cite{Socher11} exploit structure of information using trees. They extract features from superpixels using CNN, combine nearby superpixels with same category recursively.

 As a special case of image parsing, face parsing amounts to labeling each pixel with eye, nose, mouth and so on. It is a basic task in face image analysis. Compared with general image parsing, it is simpler since facial parts are regular and highly structured. Nevertheless, it is still challenging since facial parts are deformable. For this task, landmark extraction is a common practice. But most landmark points are not well-defined and it is difficult to encode uncertainty in landmarks like nose ridge \cite{Smith13}. Segmentation-based methods seem to be more promising \cite{Smith13}\cite{Luo12}.

In the paper, we present a deep learning method for face parsing. Inspired by the models for general image parsing \cite{Farabet13}\cite{Pinheiro14}, we use multiple CNNs for processing different scales of the image. To allow the CNNs exchange information, an interlinking layer is designed, which concatenates the feature maps of neighboring CNNs in the previous layer together after downsampling or upsampling. For this reason, the proposed model is called {\it interlinked CNN} or {\it iCNN} for short. The idea of interlinking multiple CNNs is partially inspired by \cite{Seyedhosseini13} where multiple classifiers are interlinked.

Experiments on a pixel-by-pixel \cite{Smith13} labeling version of the Helen dataset \cite{Le12} demonstrate the effectiveness of iCNN compared with existing models.

\section{iCNN}

\begin{sidewaysfigure}
\centering
\includegraphics[width = 0.95\textwidth]{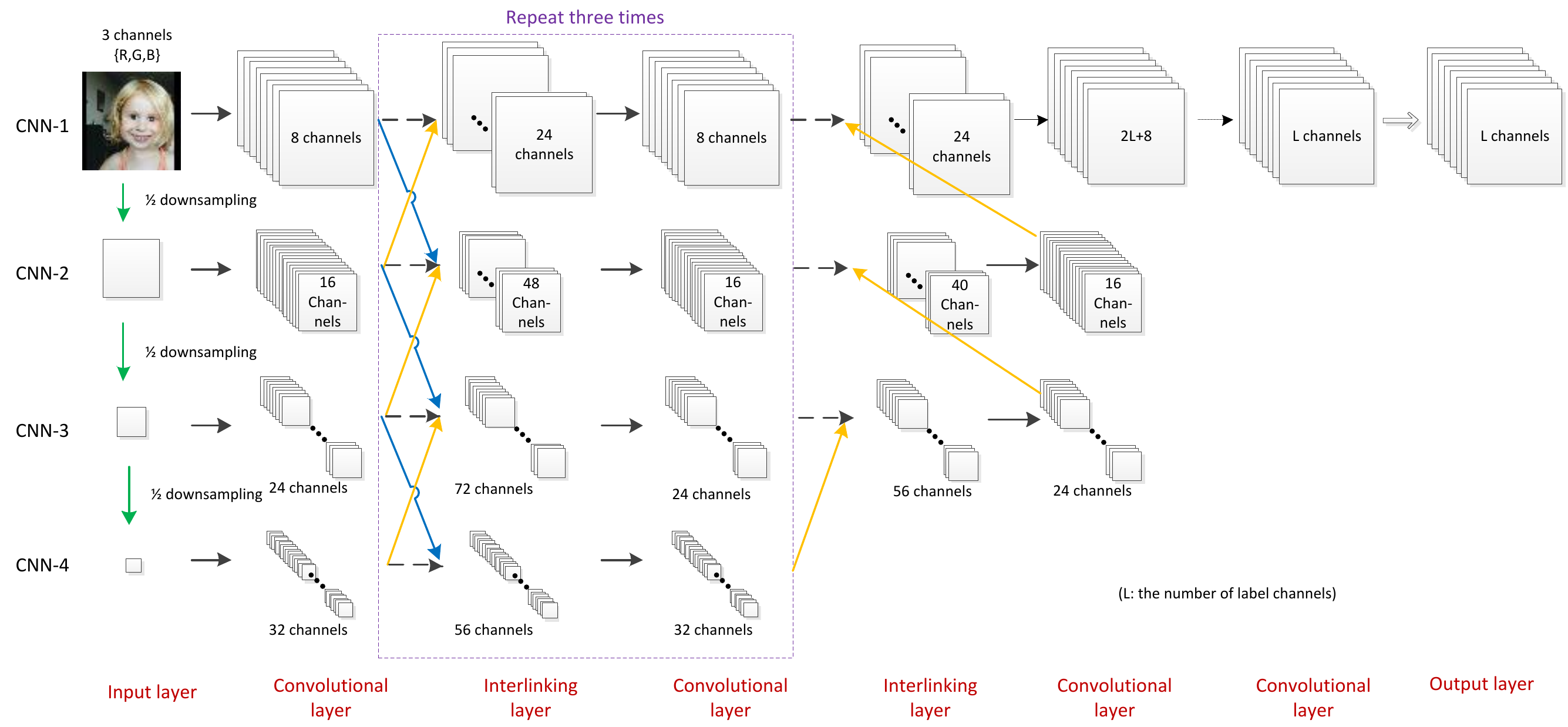}
\caption{The structure of the iCNN. Black solid arrows: convolution and nonlinear transformation as described in \eqref{E:conv}. Black dotted arrow: convolution. Black open arrow: softmax. Green arrows: downsampling (mean pooling). Black dashed arrows: pass the feature maps to the next layer. Blue arrows: downsampling (max pooling) the feature maps and pass them to the next layer. Yellow arrows: upsampling (nearest neighbor) the feature maps and pass them to the next layer. Best viewed in color.}
\label{fig:iCNN}
\end{sidewaysfigure}

The overall structure of the proposed iCNN is illustrated in Fig. \ref{fig:iCNN}. Roughly speaking, it consists of several traditional CNN in parallel, which accept input in different scales, respectively. These CNNs are labeled CNN-1, CNN-2, ... in the order of decreasing scale. The hallmark of the iCNN is that the parallel CNNs interact with each other. From left to right in Fig. \ref{fig:iCNN}, the iCNN consists of alternating convolutional layers and interlinking layers, as well as an output layer, which are described as follows.

\subsection{Convolutional Layers}
The convolutional layers are the same as in the traditional CNN, where local connections and weight sharing are used. For a weight kernel $w_{uvkq}^{(l)}$, the output of a unit at $(i,j)$ in the $l$-th layer is
\begin{equation}\label{E:conv}
y_{ijq}^{(l)}=f\left(\sum_{k=1}^{C}\sum_{u=1}^{P_1}\sum_{v=1}^{P_2}
w_{uvkq}^{(l)} y_{i+u,j+v,k}^{(l-1)}+b^{(l)}\right)
\end{equation}
where $P_1$ and $P_2$ denote the size of the weight kernel in the feature map, $C$ denotes the number of channels in the  $(l-1)$-th layer, $b^{(l)}$ denotes the bias in the $l$-th layer, and $f(\cdot)$ is the activation function. Throughout the paper, tanh function is used as the activation function. If we use $Q$ kernels $w_{uvkq}^{(l)}$, that is, $q=1, \ldots, Q$, then a total number of $Q$ feature maps (the $q$-th feature map consists of $y_{ijq}^{(l)}$ for all $i,j$) will be obtained in  the $l$-th layer.

The operation in the bracket in \eqref{E:conv} can be implemented by tensor convolution. The surrounding of feature maps in the  $(l-1)$-th layer are padded with zeroes such that after convolution and activation the resulting feature maps in the $l$-th layer has the same size in the first two dimensions as the feature maps in the  $(l-1)$-th layer.

\subsection{Interlinking Layers}

In conventional CNN \cite{LeCun98}\cite{Krizhevsky12}, there are downsampling layers which perform local max pooling or average pooling. They can realize shift invariance, which is important for pattern recognition. Downsampling reduces the size of feature maps. This is not a problem for pattern recognition (instead it is preferred because it reduces the computational burden in subsequent layers), but becomes problematic for scene parsing if an end-to-end model is desired. The output of an end-to-end model should have the same size as the input image in the first two dimensions because we have to label every pixel. Considering this requirement, we do not perform downsampling in the first CNN (top row in Fig. \ref{fig:iCNN}). The other CNNs (other rows in Fig. \ref{fig:iCNN}) process the input in smaller scales, and we do not perform downsampling in their own previous feature maps, either (black dashed arrows in Fig. \ref{fig:iCNN}).

These parallel CNNs process different scales of the input, which contain different levels of fine to coarse information. To let each CNN utilize multi-scale information, a special layer is designed. Consider CNN-$k$. In this layer, the feature maps from its own previous layer and those from the previous layer of CNN-$(k-1)$ and CNN-$(k+1)$ are concatenated. But the three types of feature maps cannot be concatenated directly because they have different sizes in the first two dimensions: those from CNN-$(k-1)$ are larger than those from CNN-$k$ and those from CNN-$(k+1)$ are smaller than those from CNN-$k$. Our strategy is to downsample those from CNN-$(k-1)$ and upsample those from CNN-$(k+1)$ such that they have the same size as those from CNN-$k$ in the first two dimensions. Max pooling is used for downsampling and nearest neighbor interpolation is used for upsampling. By performing downsampling/upsampling and then concatenation, we have interlinked the parallel CNNs.

\subsection{Output Integration}

It has been seen that after either the convolutional layer or interlinking layer, the size of the feature maps of each CNN in the first two dimensions do not change. Only CNN-1's feature maps have the same size as the output tensor in the first two dimensions. To utilize the information of other CNNs, we perform the following steps for $k=4,3,2$ in sequel:
\begin{enumerate}
\item upsample CNN-$k$'s final feature maps to match the size of CNN-$(k-1)$'s feature maps in the first two dimensions,
\item concatenate these feature maps with those from CNN-$(k-1)$, and
\item perform convolution and nonlinear transformation using \eqref{E:conv} to obtain a bunch of CNN-$(k-1)$'s final feature maps.
\end{enumerate}
After these operations, an additional convolutional layer without nonlinear transformation is used in CNN-1 with $L$ feature maps, where $L$ denotes the number of different labels. See Fig. \ref{fig:iCNN} for illustration.

\subsection{Output Layer}

Only CNN-1 has a softmax layer in the end, which output the labels of each pixel. The output is a 3D tensor with the first two dimensions corresponding to the input image and the third dimension corresponding to the labels. At each location of the pixel, the one-hot representation is used for labels, that is, there is only one element equal to one and others equal to zero along the third dimension.

\subsection{Training}

The cross-entropy function is used as the loss function. Same as other CNNs, any minimization technique can be used. Stochastic gradient descend is used in this project.

\subsection{Parameter Setting}

For this particular application, the input image has a size of either  $64\times 64$ or $80\times 80$. There are two stages in the proposed face parsing pipeline where in the first stage the entire image is resized (downsampling) to $64\times 64$ and in the second stage $64\times 64$ and $80\times 80$ patches are extracted in the original image to cover the eye/nose/eyebrow and the mouth, respectively. See the next section for details. For RGB images, the input has three channels. The input image is then downsampled to 1/2, 1/4 and 1/8 size using a $2\times 2$ mean pooling. In all convolutional layers and all CNNs, the size of the receptive field is set to $5\times 5$ (the first two dimensions) except in the last convolutional layer of CNN-1 (the black dotted arrow) where $9\times 9$ is used.

\section{Face Parsing with iCNNs}

Usually a face image for parsing is large, e.g., the images of Helen dataset \cite{Le12} for this task are of the size $256\times 256$ \cite{Smith13}. If we input such large images to the proposed iCNN, both training and testing are slow. To speed up the process we separate the face parsing procedure into two stages, and both stages use iCNN.

\subsection{Stage 1: Facial Parts Localization}\label{subsec:loc}

\begin{figure}
\centering
\includegraphics[width = \textwidth]{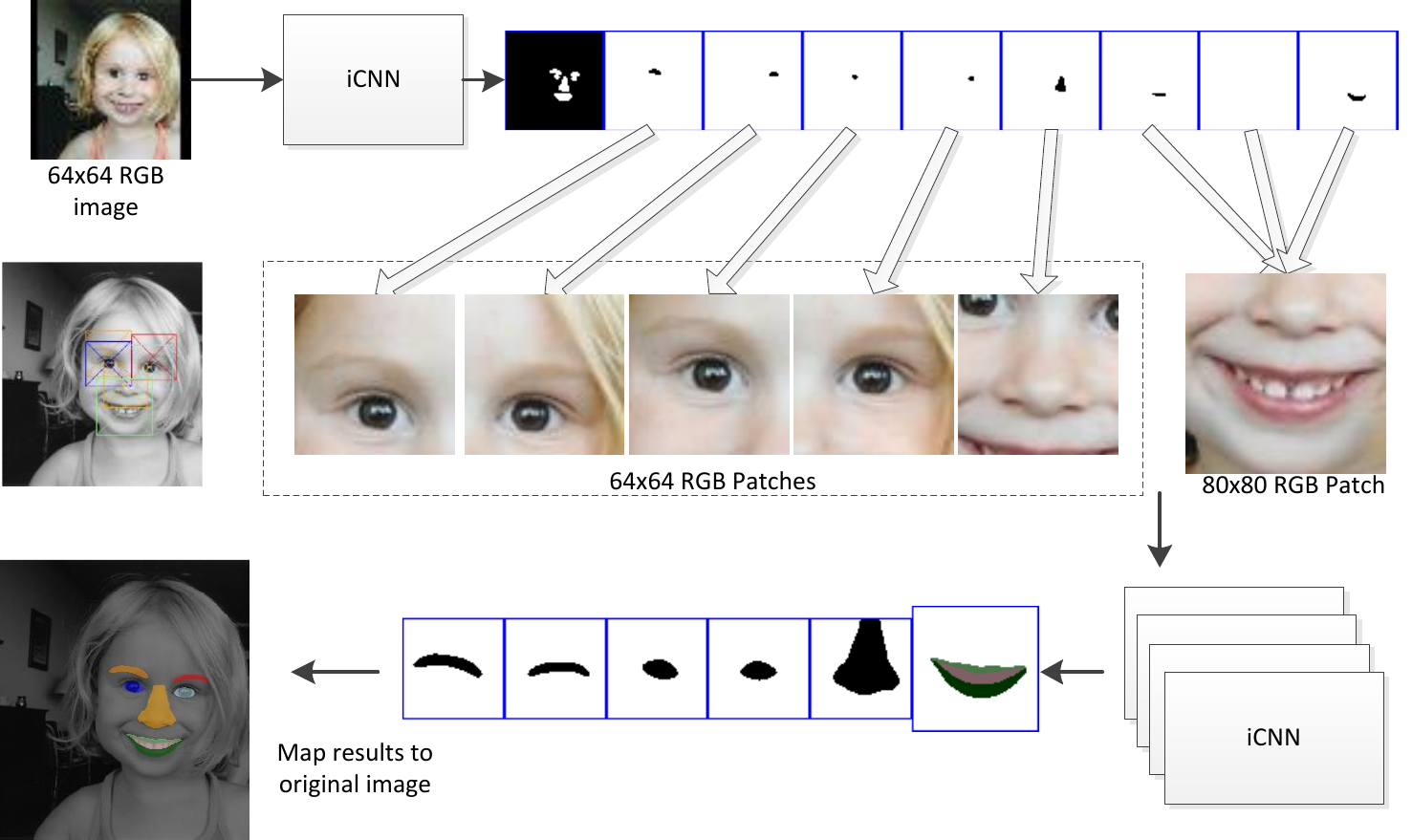}
\caption{The pipeline for face parsing. In the first stage, the entire image is resized to $64\times 64$ with aspect ratio kept. It is input to an iCNN and obtain ten label maps where the first is the background and the others are facial components. The median point of each component except the two lips and in-mouth is calculated. Since the lips and in-mouth are processed together, these three parts are first merged together and then a joint median point is calculated. A $64\times 64$ or $80\times 80$ patch is extracted around the median point. In the second stage, with mirroring operation (right eye and eyebrows flipped), six small parts are processed by four iCNNs to get exact segmentation at the pixel level. Best viewed in color.}
\label{fig:pipeline}
\end{figure}

The goal of this stage is to localize the facial parts including the eyes, nose and so on with iCNN. Note that we do not label the Face Skin part in this project, since it has a large area, which is unsuitable for the proposed iCNN to process. The input image is preprocessed by subtracting the mean and dividing the norm. The input image as well as its label map is resized to $64\times 64$ in the first two dimensions (both the input image and the output map are 3D tensors) by downsampling. The output tensor has 9 channels corresponding to the label maps of background, left eyebrow, left eye, right eyebrow, right eye, nose, upper lip, inner mouth and lower lip, respectively (Fig. \ref{fig:pipeline}). Except the first label map (background), each median axis of the label map is calculated, and scaled back to original image to obtain the estimation of the part location. For mouth related parts (upper lips, inner mouth, lower lips), a shared median axis is calculated. For the first five parts, $64\times 64$ patches are extracted from the original input face image. For mouth-related parts, a $80\times 80$ patch is extracted.

\subsection{Stage 2: Fine Labeling}

In the previous stage, we have extracted the five $64\times 64$ patches and one $80\times 80$ patch from the original image. Then we use four iCNNs to predict the labels of the pixels in each patch (Fig. \ref{fig:pipeline}). The four iCNNs are used for predicting eyebrows, eyes, nose, and mouth components, respectively. Note that one iCNN is used for predicting both left eyebrow and right eyebrow. Since the left eyebrow and right eyebrow are symmetric, during training the image patches of right eyebrows are flipped and combined with image patches of left eyebrows. Therefore this iCNN has only one label map in the output. In testing, the predicted label maps of right eyebrows are flipped back. Similarly, one iCNN is used for predicting both left eye and right eye. The iCNN for the nose has only one label map in the output and the iCNN for the mouth components has three label maps.

\section{Experiments}

\subsection{Dataset}

\begin{figure}
\centering
\includegraphics[width = 0.7\textwidth]{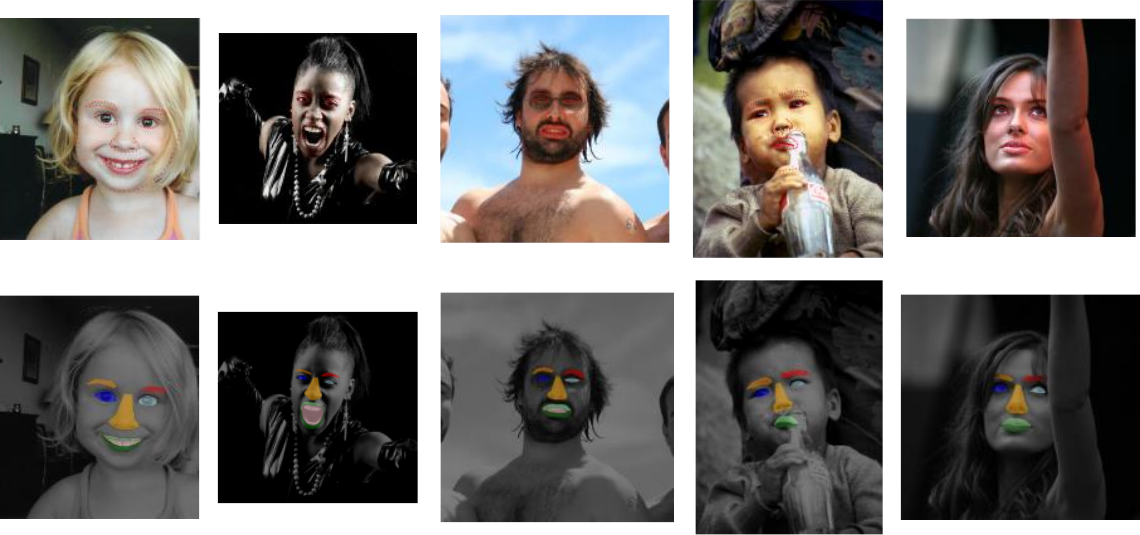}
\caption{Example images (top) and corresponding labels (bottom) of the Helen dataset. Best viewed in color.}
\label{fig:helen}
\end{figure}

The Helen dataset \cite{Le12} is used for evaluation of the proposed model, which has 2330 face images with dense sampled, manually-annotated contours around the eyes, eyebrows, nose, outer lips, inner lips and jawline. It is originally designed as a landmark detection benchmark database. Smith et al. \cite{Smith13} provides a resized and roughly aligned pixel-level ground truth data to benchmark the face parsing problem. It generates ground truth eye, eyebrow, nose, inside mouth, upper lip and lower lip segments automatically by using the manually-annotated contours as segment boundaries. Some examples of Helen are shown in Fig. \ref{fig:helen}, where the first line is the original database images with annotations, and second line is the processed pixel-based labeling for parsing.

We use the same training, testing and validation partition as in \cite{Smith13}. The dense annotated data is separated into 3 parts:  2000 images for training, 230 images for validation, and 100 images for testing. The validation set is used to test whether model is converged.

\subsection{Training and Testing}

We train the iCNNs in stages 1 and 2 separately. For stage 1, the entire training images, as well as the corresponding ground truth label maps, are resized to $64\times 64$ with aspect ratio kept. For stage 2, the training data are $64\times 64$ or $80\times 80$ patches extracted from the original $256\times 256$ training images (see Section \ref{subsec:loc}). The corresponding ground truth label maps are extracted from the original $256\times 256$ ground truth label maps.

Stochastic gradient descent is used as the training algorithm. Since the number of images is small compared to number of parameters, to prevent overfitting and enhance our model, data argumentation is used. During stochastic gradient descent, a random $15^\circ$ rotation, $0.9\scriptsize{\sim}1.1$x scaling, and $-10\scriptsize{\sim}10$ pixels shifting in each direction are applied to each input every time when it enters the model.

In Stage 2, by visualizing the feature maps, we find that in the last convolutional layer of CNN-1 among the $L$ feature maps there is a feature map, denote dy $B$, corresponding to the background part. We find that modulating this feature map by $\beta B+\beta_0$ can enhance the prediction accuracy. For each facial part, $\beta$ and $\beta_0$ are obtained by maximizing the F-measure \cite{Smith13} on the validation set using the L-BFGS-B algorithm offered by SciPy, an open-source software.

For testing, each image undergoes stages 1 and 2 in sequel. Only the predicted labels in stage 2 are used for evaluation of the results.

All codes are written in Theano \cite{Bergstra10} and Pylearn2 \cite{Goodfellow13}.

\subsection{Results}

\begin{figure}
\centering
\includegraphics[width = 0.8\textwidth]{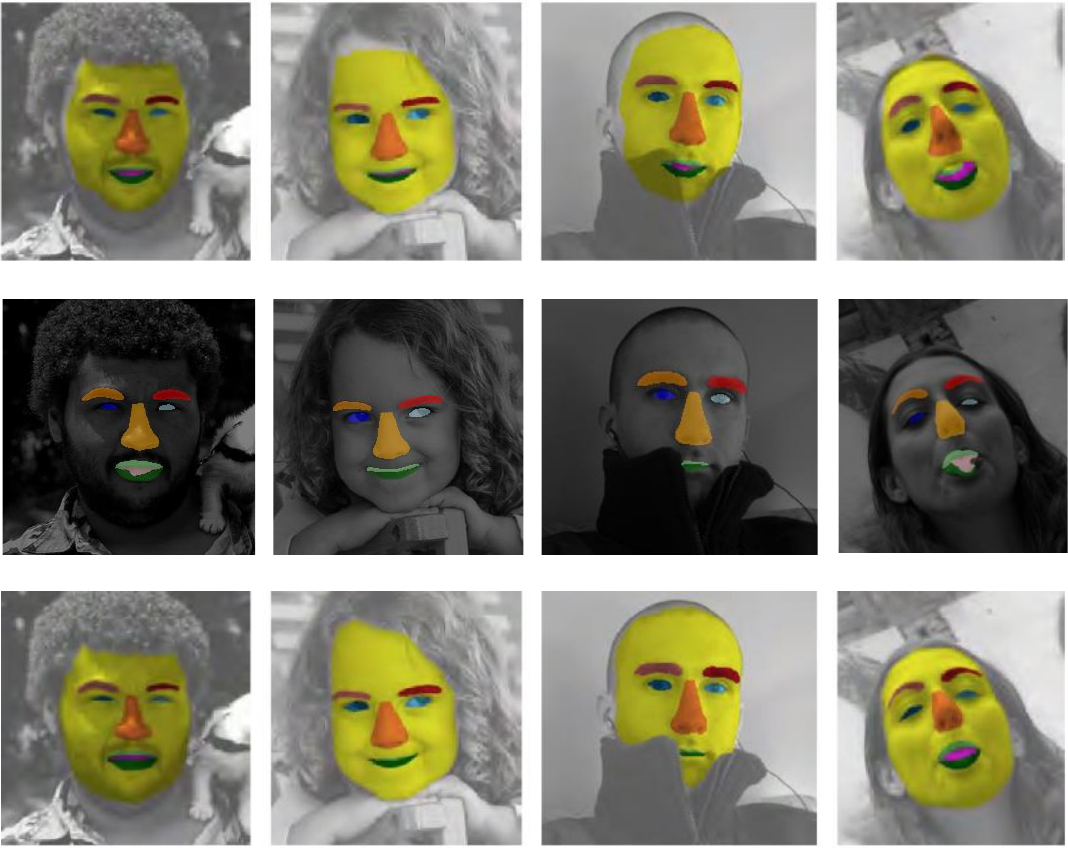}
\caption{Labeling results on several example images obtained using the method in \cite{Smith13} (top) and the proposed method in this paper (middle). The bottom shows the ground truth labels. Best viewed in color.}
\label{fig:result}
\end{figure}

The  evaluation metric is the F-measure used in \cite{Smith13}. From Table \ref{tab:comparison}\footnote{The results of iCNN are incorrect in our ISNN2015 paper}, it is seen that for most facial parts, iCNNs obtain the highest scores. Note that in our training data, the labels of Face Skin area are not used. As we can see in the table, this area is usually a high-score term for most methods, and omitting it will in no way enhance the overall performance of iCNNs. Even though, iCNNs achieves higher overall score than existing models. Some example labeling results are shown in Fig. \ref{fig:result} along with the results obtained in \cite{Smith13}.

\begin{table}
\caption{Comparison with other models (F-Measure)}
\begin{tabular}{c|c|c|c|c|c|c|c|c|c}
  \hline
  Model	&Eye&	Eyebrow&	Nose&	In mouth&	Upper lip&	Lower lip&	Mouth (all)&	Face Skin&	Overall\\
  \hline\hline
\cite{Zhu12}&	0.533&	n/a&	n/a&	0.425&	0.472&	0.455&	0.687&	n/a&	n/a\\
\cite{Saragih09}&	0.679&	0.598&	0.890&	0.600&	0.579&	0.579&	0.769&	n/a&	0.733\\
\cite{Liu11}&	0.770&	0.640&	0.843&	0.601&	0.650&	0.618&	0.742&	0.886&	0.738\\
\cite{Gu08}&	0.743&	0.681&	0.889&	0.545&	0.568&	0.599&	0.789&	n/a&	0.746\\
  \cite{Smith13}&	{\bf 0.785}&	0.722&	{\bf 0.922}&	0.713&	0.651&	0.700&	0.857&	0.882&	0.804\\
\hline
iCNNs&	0.778&	{\bf0.863}&	0.920&	{\bf0.777}&	{\bf0.824}&	{\bf0.808}&	{\bf0.889}&	n/a&	{\bf 0.845}\\
  \hline
\end{tabular}
\label{tab:comparison}
\end{table}

\section{Concluding Remarks}

We propose an interlinked CNN (iCNN), where multiple CNNs process different levels of details of the input, respectively. Compared with traditional CNNs it features interlinked layers which not only allow the information flow from fine level to coarse level but also allow the information flow from coarse level to flow to the fine level. For face parsing, a two-stage pipeline is designed based on the proposed iCNN. In the first stage an iCNN is used for facial part localization, and in the second stage four iCNN are used for pixel labeling. The pipeline does not involve any feature extraction step and can predict labels from raw pixels. Experimental results have validated the effectiveness of the proposed method.

Though this paper focuses on face parsing, the proposed iCNN is not restricted to this particular application. It may be useful for other computer vision applications such as general image parsing and object detection.

\section*{Acknowledgement}

The first author would like to thank Megvii Inc. for providing the computing facilities. This work was supported in part by the National Basic Research Program (973 Program) of China under Grant 2012CB316301 and Grant 2013CB329403, in part by the National Natural Science Foundation of China under Grant 61273023, Grant 91420201, and Grant 61332007, in part by the Natural Science Foundation of Beijing under Grant 4132046.

\end{document}